%% file: main.tex
\documentclass[a4paper,conference]{IEEEtran}

\usepackage{times}
\usepackage{amsmath}
\usepackage{amssymb}
\usepackage{dblfloatfix}
\usepackage{comment}
\usepackage{float}

\usepackage[T1]{fontenc}
\usepackage[utf8]{inputenc}
\usepackage[american]{babel}

\usepackage{multirow}

\usepackage{epsfig}
\usepackage[dvipsnames,svgnames,x11names,table, dvipsnames]{xcolor}
\usepackage[export]{adjustbox}
\graphicspath{{images/}}

\input{macros}

\begin{document}

\setlength{\parskip}{0.0pt plus 0pt minus 0.55pt}
%
\title{Modeling Long-Term Interactions to Enhance Action Recognition}

\author{\IEEEauthorblockN{Alejandro Cartas}
\IEEEauthorblockA{University of Barcelona\\
alejandro.cartas@ub.edu}
\and
\IEEEauthorblockN{Petia Radeva}
\IEEEauthorblockA{University of Barcelona\\
petia.ivanova@ub.edu}
\and
\IEEEauthorblockN{Mariella Dimiccoli}
\IEEEauthorblockA{Institut de Robòtica i Informàtica Industrial (CSIC-UPC)\\
mdimiccoli@iri.upc.edu}}
\thanks{This work was partially supported by the Spanish Ministry of Economy and Competitiveness and the European Regional Development Fund (MINECO/ERDF, EU) through the program Ramon y Cajal, projects PID2019-110977GA-I00 and RED2018-102511-T. We acknowledge the support of NVIDIA Corporation with the donation of Titan Xp GPUs.}


\maketitle

\begin{abstract}

In this paper, we propose a new approach to understand actions in egocentric videos that exploits the semantics of object interactions at both frame and temporal levels. At the frame level, we use a region-based approach that takes as input a primary region roughly corresponding to the user hands and a set of secondary regions potentially corresponding to the interacting objects and calculates the action score through a CNN formulation. This information is then fed to a Hierarchical Long Short-Term Memory Network (HLSTM) that captures temporal dependencies between actions within and across shots. Ablation studies thoroughly validate the proposed approach, showing in particular that both levels of the HLSTM architecture contribute to performance improvement. Furthermore, quantitative comparisons show that the proposed approach outperforms the state-of-the-art in terms of action recognition on standard benchmarks, without relying on motion information.

\end{abstract}


%
\IEEEpeerreviewmaketitle

\section{Introduction}

Activity recognition has been a very active research area in computer vision during the last decade \cite{poppe2010survey}. 
In recent years, progress in wearable camera technology has offered the opportunity to capture naturally-occurring activities from a first-person point of view. The egocentric paradigm is particularly beneficial for analyzing activities involving object manipulations \cite{dimiccoli2018computer}, commonly referred to as {\em human-object interactions}.
Following the egocentric literature on the latter \cite{fathi2011understanding,Ma_2016_CVPR}, the term \textit{action} represents a short-term human-object interaction such as \textit{open ketchup} or \textit{close fridge}, consisting of an action verb (i.e. open) and an interacting object (i.e. ketchup). The term \textit{activity} indicates a sequence of actions such as {\em take bread, take cheese, open ketchup,} etc. performed with a specific goal, i.e. making a sandwich. 
Most works about human-to-object interaction model actions through egocentric features such as hand location and pose \cite{bambach2015lending,fathi2011understanding}, head motion and gaze \cite{li2015delving}, manipulated objects \cite{mccandless2013object,pirsiavash2012detecting,Baradel_2018_ECCV}, or as a combination of all them \cite{Ma_2016_CVPR,behera2012egocentric,sudhakaran2018}. However, only a few of them \cite{fathi2011understanding,Baradel_2018_ECCV,sudhakaran2018} specifically aimed at reasoning about different semantic entities in images across space and/or time.

In the egocentric setting, the temporal structure context has been used for temporal segmentation \cite{poleg2014temporal}, story-telling \cite {gupta2009, Xiong2015}, and recognizing human interaction activities \cite{Ryoo2016}. Only a few works have been focused on exploiting the temporal order of actions. \cite{fathi2011understanding} modeled contextual relations between actions and activities and the ordering between adjacent actions. In \cite{Baradel_2018_ECCV}, the spatio-temporal reasoning is performed on an object-level by relying on state-of-the-art object segmentation networks that allow tracking the object appearance over time.

\color{Black}

\begin{figure}[!t]
\begin{center}
\includegraphics[width=0.9\columnwidth]{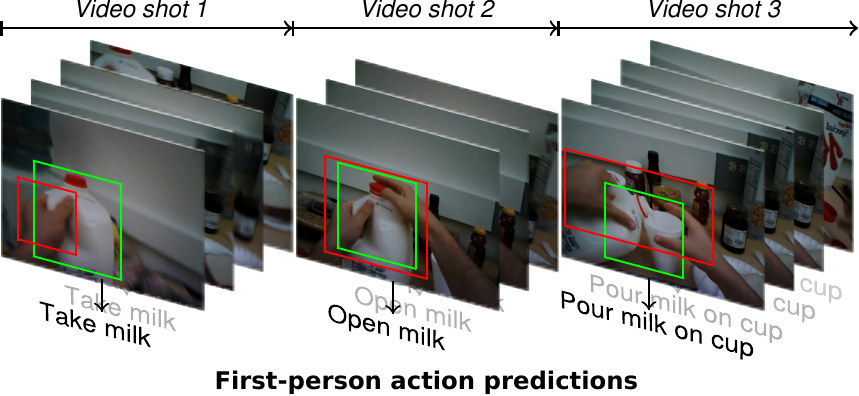}
\vspace{-2mm}
\caption[]{Our model takes as input for each frame a primary region (user hands, \textcolor{red}{red} boxes) and a set of secondary regions (object proposals, \textcolor{green}{green} boxes). These regions are used to make action predictions at frame level, that are aggregated firstly at shot level and then across shots.}\label{fig:actionSequences}
\end{center}
\end{figure}


\begin{figure*}[t!]
\begin{minipage}{\textwidth}%
\includegraphics[width=\textwidth]{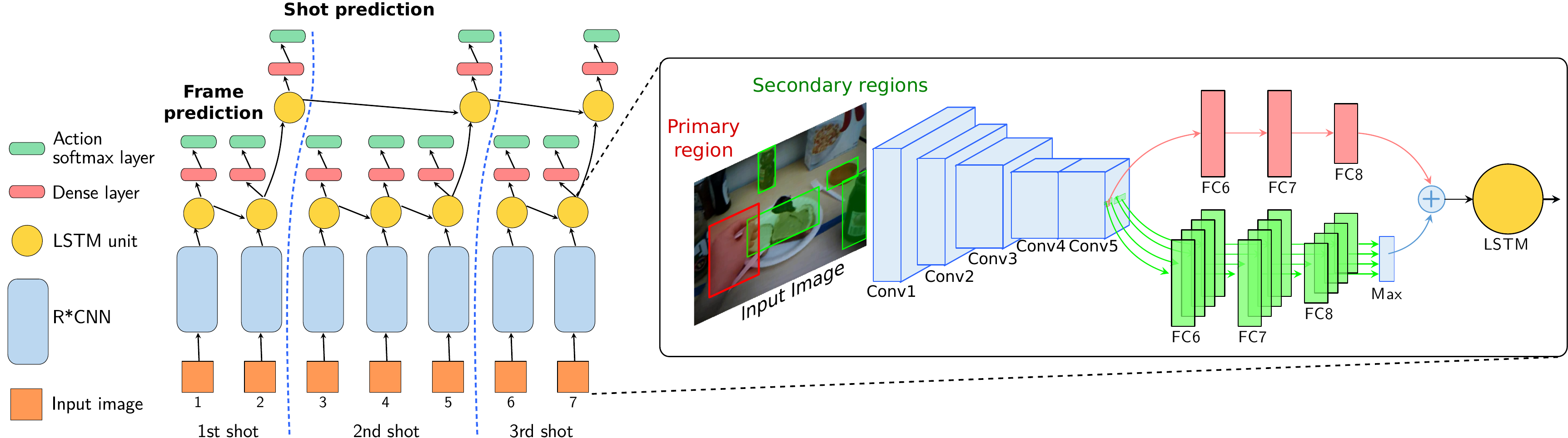}%
\end{minipage}%
\caption{Overview of our proposed architecture. The left part shows the hierarchical action recognition pipeline across different video shots. The output of the last LSTM unit of each shot in the first level serves as input for an LSTM unit of the second level of the hierarchy. The right part shows the frame level processing, the primary (\textcolor{red}{red}), and secondary regions (\textcolor{green}{green}) boxes are previously estimated and passed as input to the network. The R*CNN architecture is applied to each frame within the video shot. The action prediction is measured at the frame level, as the output of the softmax of the first layer of LSTM.}
\label{fig:architecture}
\end{figure*}

In this work, we present a new egocentric action recognition approach that is able to reason about relevant image regions at the frame level, and also to understand action dynamics at the temporal level. At the frame level, we model first-person actions as spatially structured regions in a deformable star configuration, consisting 
of primary and secondary regions corresponding to the user hands and to the objects being potentially interacted with, respectively (see Fig. \ref{fig:actionSequences}). We adapt the CNN framework proposed in \cite{rstarcnn2015} to train jointly the egocentric action-specific models and the feature maps.
At the temporal level, we model first-person activities as sequences of actions and we propose a Hierarchical LSTM architecture, whose first layer captures the temporal evolution of contextual features over time within shots and the second one captures long-term temporal dependencies between actions across shots. 

We evaluated the proposed approach on the GTEA, GAZE, and GAZE+ benchmarks. Quantitative comparisons to five state-of-the-art methods demonstrated both the effectiveness and the potential of our approach, which is able to outperform current methods even without relying on motion information. Unlike most available architectures, where motion is processed through a dedicated stream, our approach explicitly models the dynamics of human-object interaction at temporal level. 

Our experiments demonstrate that effectively exploiting semantics can have a major impact on performance than exploiting modeling motion. Furthermore, with this work, we provide useful insights on (1) the role of spatial reasoning at image level and (2) the role of reasoning over the temporal structure of human-object interactions.


\input{tabStateArt}

\section{Related work}
\label{sec:StateOfTheArt}

\paragraph{Third-person action recognition from videos}

Comprehensive surveys that cover the pre-deep learning era, can be found in \cite{poppe2010survey,weinland2011survey}. In recent years, there has been a proliferation of approaches based on deep learning architectures. Most of these works \cite{simonyan2014two,donahue2015,ng2015,peng2016multi} focus on atomic actions such as \textit{brushing hairs}, \textit{shake hands}, \textit{kiss}, etc., instead of a sequence of actions aiming at performing a given activity. Typical neural network architectures for action recognition include ConvNet+LSTM \cite{donahue2015,ng2015}, two-stream convolutional networks \cite{Carreira_2017_CVPR,peng2016multi,simonyan2014two}, 3D ConvNet \cite{ji20133d} and Long Temporal Convolutions \cite{varol2018long}. In ConvNet-LSTM architecture, LSTM cells are typically added at the end of a CNN. Furthermore, the network input is a sequence of frames sampled from a video. In two-stream convolutional networks, one of the streams captures the spatial information while the other carries the temporal information. 3D ConvNets are a natural extension of 2D CNNs to video that allows us to capture invariance to translation in time. Finally, in LTC 3D, ConvNets are further extended to much longer temporal convolutions that enable action representation at their full-scale \cite{varol2018long}.

\paragraph{First-person action recognition from videos}

Early works were focused on modeling spatio-temporal features. \cite{fathi2011understanding} modeled activities as a
sequence of actions and each action as a spatio-temporal relationship between the hands and the objects involved in it. This work was further extended to include gaze features \cite{fathi2012learning}. \cite{pirsiavash2012detecting} introduced a temporal pyramid approach to encode spatio-temporal features along with detected active objects.\cite{mccandless2013object} proposed a randomized spatio-temporal pyramids framework able to model the frequency with which each object category appears in particular space-time regions.

More recently, several end-to-end learning approaches have been proposed. \cite{bambach2015lending} focused on hand location and pose, \cite{li2015delving} on object features and gaze. Ma et al. \cite{Ma_2016_CVPR} proposed a twin stream RGB-optical flow CNN architecture. Singh et al. \cite{Singh_2016_CVPR} proposed four-stream neural network architecture. The first two streams capture temporal features from binary hand-arm masks. While the other two process spatial information from an RGB frame and motion taken from stacked optical flow encoded in a saliency map, respectively. \cite{sudhakaran2018} proposed an attention mechanism that enables the network to attend regions containing objects correlated with the activity under consideration. The output of the attention module serves as input for an LSTM network. \cite{Baradel_2018_ECCV} proposed a network able to reason about detected semantic object instances through space and time, building on the result of an object detector.
 
Differently to \cite{Baradel_2018_ECCV,sudhakaran2018} who fed to the recurrent units the features descriptions of relevant objects, we first perform action prediction by reasoning at the frame level, and then we work with action probabilities at the temporal level. Our temporal model is conceptually similar to \cite{fathi2011understanding}, but contrary to them, we do not rely on domain knowledge about the activity being performed in the prediction process. In Table \ref{tab:state-of-the-art}, we list current state-of-the-art approaches to first-person action recognition along with ours and indicate the characteristics of each one in terms of features used and type of modeling.

\paragraph{Temporal context for action recognition}

The role of the temporal context in action recognition has been the focus of works on action anticipation \cite{Yuge18} and detection \cite{sstad_buch_bmvc17}. The former task consists in recognizing an action from a video shot as early as possible. The latter task determines the set of actions occurring in long videos and their time boundaries, but the actions do not necessarily belong to a single activity. The idea of using the temporal context for improving action recognition was explored in \cite{Bobick1998,fathi2011understanding}. Based on the observation that adding knowledge of the temporal structure of events helps to disambiguate low-level features on sequences of actions, \cite{Bobick1998} proposed a stochastic grammar parsing over the results of a Hidden Markov Model. 

\section{Proposed approach}
\label{sec:approach}



\subsection{Frame-level modelling}
\label{subsec:mod_context}

To capture contextual cues of actions, we employed a star-structured region model with CNN features, adapting the R*CNN framework proposed in \cite{rstarcnn2015} for action recognition in still images captured by a third-person camera. This approach takes as input a manually selected primary region containing the person whose action has to be predicted, and a set of secondary regions that are supposed to capture contextual information such as objects the person is interacting with, the pose of the person or what other people in the image are doing. 
In our egocentric setting, the model is defined by a previously detected primary region corresponding to hands plus a set of parts regions corresponding to manipulated objects with associated CNN features.

\begin{figure*}[!t]
\begin{minipage}[t]{\textwidth}
\begin{center}
\includegraphics[width=0.85\textwidth]{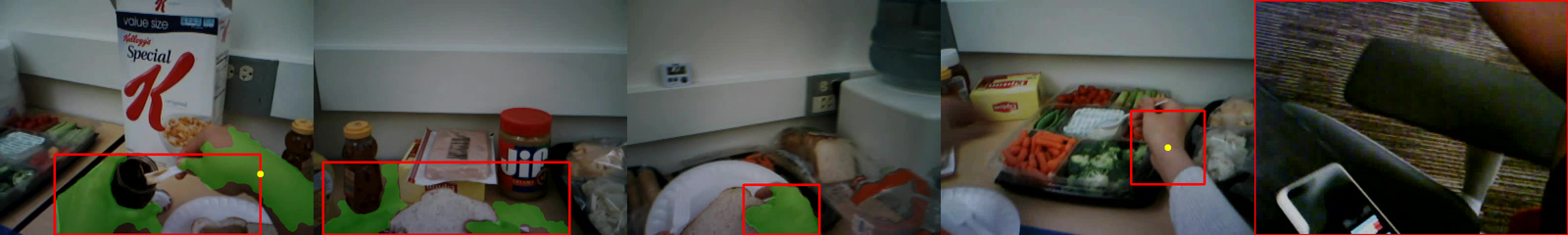}
\end{center}
\end{minipage}%
\begin{center}
\caption[]{Examples of primary regions extraction taken from random frames of the GAZE dataset. The green-colored regions show detected skin pixels, the yellow dots are the wrist center locations, and the red bounding box shows the located primary region.}
\label{fig:primaryRegionExtraction}
\end{center}
\end{figure*}

\paragraph{Primary region detection}
\label{subsec:primaryRegionDetection}

Since the human-to-object interactions are mainly done by manipulating the objects with the hands, we focus on the area covering the hand starting from the wrist to the tip of its middle finger. We define a primary region as the single area covered by one or both hands of the person. Fig. \ref{fig:actionSequences} depicts three examples where the primary regions detected with the proposed approach are shown in red bounding boxes.
Our method determines the primary region based on skin regions and wrists located points. 
First, we perform skin pixel detection using the algorithm proposed in \cite{Li_2013_ICCV_Workshops}. The training examples for this method are images containing visible arms and their respective skin masks. This method trains a random forest for each training example using a cluster from a set of local appearance features, such as different colorspaces (RGB, HSV, LAB) or spatial descriptors (SIFT \cite{Lowe2004}, HOG \cite{Dalal2005}, BRIEF \cite{Calonder2010}). The skin classification is done by globally combining the output of each random forest using a probabilistic model. In this work, the skin pixel detector was trained using only colorspaces.

The location of the wrists is performed using the Parts Affinity Fields model \cite{cao2017realtime}. This CNN model has two branches that separately predict confidence maps for body part detection and part affinity fields (PAFs). The PAFs encode the orientation and location of limbs over the image. Both kinds of confidence maps are associated using a bipartite graph matching algorithm. The result are body poses for each person on the scene. Specifically, we only used the confidence maps generated for body part detection, since other body joints are not visible and their association cannot be made.

Once the skin regions and wrist location points were calculated, the primary region is determined as follows. If no skin regions and wrist location points are found, then the primary region is considered to be the full-frame (see the last column in the second row of Fig. \ref{fig:primaryRegionExtraction}). When no wrist points are located, then the complete skin region is determined to be the primary region. This case usually happens when a small part of the hand is seen (see the third column on the first row Fig. \ref{fig:primaryRegionExtraction}). In case, where no skin regions are found, then a fixed area near the wrists is considered as primary region (see the fourth column of the second row on Fig. \ref{fig:primaryRegionExtraction}). Finally, when both wrists points are located and skin regions are found, then wrists points determine the right, left and lower sides of the primary region, while the skin determine its upper side (i.e. first two columns of the first row of Fig. \ref{fig:primaryRegionExtraction}).

\paragraph{Secondary regions prediction}

In contrast to \cite{rstarcnn2015}, where secondary regions are defined as those that overlap the primary region, our set of secondary regions coincides with the set of proposed regions. The reason is that often while performing action as \textit{take}, \textit{poor water} etc., the hands are not in contact with the object during the full duration of the shot. Considering the performance results obtained in \cite{Hosang2014Bmvc}, we compute region proposals by using Multiscale Combinatorial Grouping (MCG) \cite{Pont2015} instead of Selective Search \cite{uijlings2013selective} as in \cite{rstarcnn2015}. 

\paragraph{Action prediction}

The score function implemented by the R*CNN architecture corresponds to those of a latent SVM formulation. In a latent SVM, each example $x$ is scored by a function of the following form $$\displaystyle f_{w}(x) = \max_{z \in Z(x)} w\cdot \phi(x,z),$$ where $w$ is a vector of model parameters, $z$ are latent values, and $\phi(x,z)$ is a feature vector.

In our particular case, given an image $x$ and the primary region $p$ in $x$ containing the hands, the score for the action $\alpha$ is defined as
\begin{equation}
\displaystyle f(\alpha; x,p) = w_p^{\alpha}\cdot\phi(x,p) + \max_{z \in Z(p,x)} w_z^{\alpha} \cdot \phi(x,z)
\end{equation} 
\noindent where $z$ is the latent variable representing a secondary region corresponding to an object, $Z(p,x)$ is the set of candidates for the secondary region (object proposals), $w_p^{\alpha}$ and $w_z^{\alpha}$ are the model parameters.


The probability that given the hand $p$ on the image $x$, and the action being performed $\alpha$, is computed using softmax as in \cite{rstarcnn2015}. 
The feature vector $\phi(\cdot)$ and the vectors of model parameters, $w_p^{\alpha}$ and $w_z^{\alpha}$, are learned jointly for all actions using a CNN in an end-to-end fashion (see Fig. \ref{fig:architecture}).

\subsection{Temporal modelling}
\label{subsec:mod_order}

Our architecture models sequences of human-object interactions using a hierarchical training structure. Variations between successive frames in a video encode useful information to make more accurate action predictions \cite{ng2015}. A standard way to account for such variations is to feed the output of a CNN to an LSTM in an end-to-end fashion. Consequently, at the bottom level of our hierarchy, we use a many-to-many LSTM to explicitly capture the temporal coherence of action probabilities within a video shot. Every LSTM unit at time step $i$ takes as inputs the action probabilities returned by R*CNN for the $i$-th frame and the hidden state from the previous LSTM unit. Additionally, each LSTM output is connected to a dense layer and a softmax to output action probabilities that account for these of previous frames (see Fig. \ref{fig:architecture}). The top-level of our hierarchy captures a long time temporal dependencies across \textit{actions}. Specifically, the LSTM output of the last frame of each shot is fed to another LSTM unit, which also takes as input the hidden state of the LSTM units of the previous shot. 
The loss function of our complete HLSTM architecture is as follows:
\begin{equation}
\mathcal{L} = (1-\beta)\mathcal{L}_N + \beta\mathcal{L}_M
\end{equation}
\noindent where $\beta$ ($0 \le \beta \le 1$) is a hyper-parameter used to trade-off between two objectives and $\mathcal{L}_N$ and $\mathcal{L}_M$ are the cross-entropy loss defined at frame-level and shot-level, respectively. They are defined as follows:
\begin{equation}
\mathcal{L}_{N} = \sum_{i=1}^N L(y^i,\hat{y}^i), \; \mathcal{L}_{M} = \sum_{j=1}^M L(y^{j},\hat{y}^{j})
\end{equation}
where $N$ is the total number of frames, $M$ is the total number of shots, $y^i$ is the ground truth action at the $i-$th frame and $\hat{y}^i$ is the action prediction, accordingly. Similarly, $y^j$ is the ground truth action at the $j-$th shot and $\hat{y}^j$ is the action prediction.

\begin{figure*}[t]
\begin{center}
\includegraphics[width=0.7\textwidth]{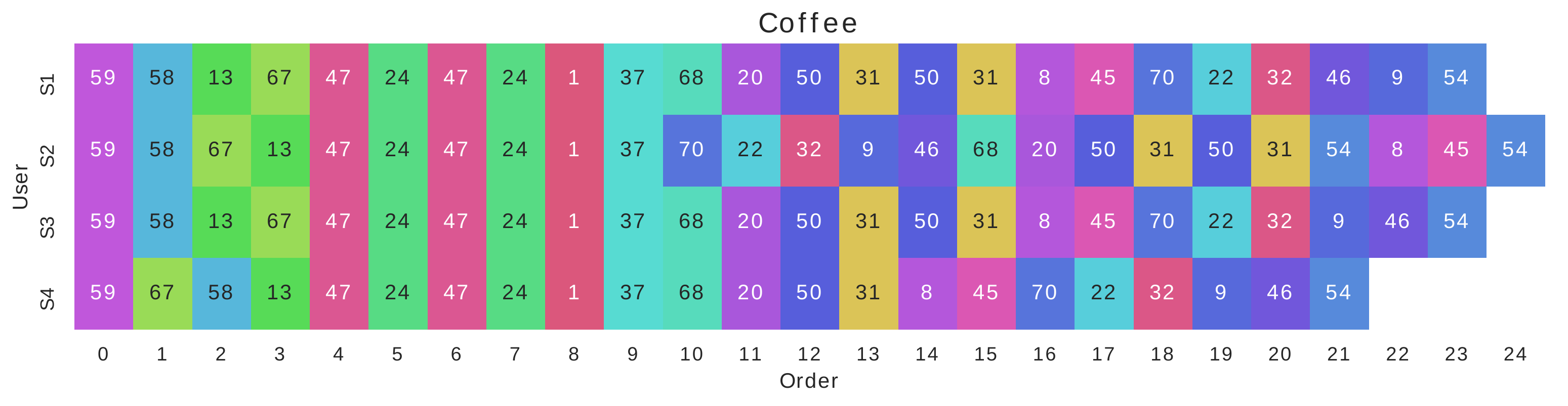}\vspace{5mm}
\includegraphics[width=0.9\textwidth]{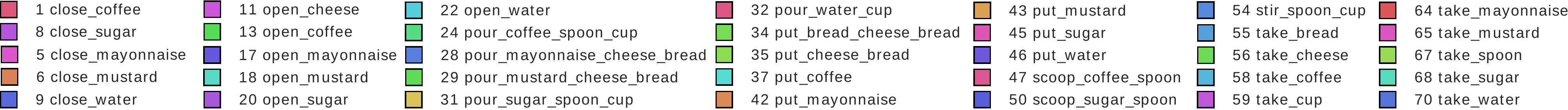}
\caption[]{Two different action sequences for all users from the GTEA dataset.}
\label{fig:gteaSequences}
\end{center}
\end{figure*}

\subsection{Data augmentation}
\label{subsec:data_augmentation}

\paragraph{Visual data augmentation}
\label{sec:visualDataAugmentation}

This data augmentation process consisted in rotating the frames of all videos in a continuous manner while maintaining the temporal consistency between consecutive ones. This tries to mimic the head movements of the camera user. Given a video from the dataset containing $N$ frames and a uniformly sampled random number $r\in\big[0,\frac{1}{2}\big]$, the rotation angle $\theta(n)$ for the $n-th$ frame is calculated using the following pair of equations:
\begin{equation}
\rho(n)= 2\pi\Big(\frac{C}{N}n+r\Big), \hspace{1mm}\theta(n) = \Theta_{max}\sum_{i}^{M}\lambda_{i}\sin{\gamma_{i}\rho(n)}
\end{equation}

\noindent where the constants $C$, $\gamma_{i}$, and $\sum_{i}^{M}\lambda_{i}=1$ determine the sinusoidal behavior of $\theta$ within a period. The purpose of the function $\theta(n)$ is to generate continuous rotation angles for each sequence in the range $[-\Theta_{max},\Theta_{max}]$.

After rotating a frame $n$ by the angle $\theta$, then its largest inscribed rectangle is cropped. This rectangle represents the augmented version of the $n-th$ frame. The width $w_r$ and height $h_r$ of this rectangle is calculated as follows. Let $s=\min(w,h)$ and $l=\max(w,h)$ denote the value of the shortest and longest side respectively.

If $\cfrac{s}{l}>2\cdot\lvert\sin{\theta}\rvert\lvert\cos{\theta}\rvert$

\begin{equation}
w_{r}=\frac{w\lvert\cos{\theta}\rvert - h\lvert\sin{\theta}\rvert}{\cos{2\theta}}, \hspace{1mm}h_{r}=\frac{h\lvert\cos{\theta}\rvert - w\lvert\sin(\theta)\rvert}{\cos{2\theta}}
\end{equation}

\noindent else if $w>h$

\begin{equation}
w_{r}=\frac{s}{2\lvert\sin{\theta}\rvert}%
,\quad%
h_{r}=\frac{s}{2\lvert\cos{\theta}\rvert}
\end{equation}

\noindent otherwise

\begin{equation}
w_{r}=\frac{s}{2\lvert\cos{\theta}\rvert}%
,\quad%
h_{r}=\frac{s}{2\lvert\sin{\theta}\rvert}
\end{equation}

The primary region previously obtained is rotated with respect to its new center location. Afterward, a new bounding box is computed using the rotated corners. The secondary regions are calculated again for the augmented version of the $n-th$ frame by using the Multiscale Combinatorial Grouping (MCG) algorithm. This process duplicated our training data.

\paragraph{Temporal data augmentation}
\label{sec:temporalDataAugmentation}

 The goal of the temporal data augmentation was to extend the sequences of actions in a logical way. For instance, the original order of three actions for a sequence named \textit{making a ham sandwich} could be \textit{take ham}, \textit{take bread}, and \textit{put ham on bread}, but it could be augmented by changing the order of the first two actions. 
 The temporal sequences of the datasets are augmented in a two-step process. First, a meta-sequence is manually created for each video sequence in the dataset. This meta-sequence defines all the possible logical combinations that the sequence can have. Second, an alternate sequence is randomly generated from its meta-sequence during training. In order to create a meta sequence, groups of related actions are created along their possible combinations. The groups can be combined using three different operations: they can be swapped/moved, skipped, or added to one another. 

\section{Experiments}
\label{sec:experimental}


\subsection{Datasets and experimental protocol}
\label{sec:datasets}
\paragraph{Datasets}

Our experiments were done using the GTEA \cite{fathi2011}, GTEA Gaze \cite{fathi2012learning}, and the GTEA GAZE+ \cite{li2015delving} datasets. The GTEA dataset consists of 21 videos acquired by 4 different subjects while performing scripted activities such as making a \textit{cheese sandwich}, or \textit{instant coffee}. Each activity contains sequences of actions that are freely performed by the subject.
The dataset includes two subsets of 71 and 61 actions. The GTEA Gaze dataset consists of 17 videos from 14 different subjects following unscripted food recipes. The dataset has two subsets of 25 and 40 actions, such as \textit{open milk} or \textit{take carrot}. Both action subsets have the same fixed training and test data splits as presented in \cite{li2015delving}. The GAZE+ dataset has 37 videos of 7 activities recorded by 6 participants. The subjects were asked to perform some activity with the available objects. The total number of evaluated actions is 40. Some examples of the activities include \textit{prepare a Greek salad} and \textit{making cheese burger}, whereas some examples of actions are \textit{put milk container} and \textit{open fridge drawer}. 
For the former three datasets, the reported results were obtained by using the same splits as in previous works \cite{fathi2011understanding,fathi2013modeling,li2015delving,Ma_2016_CVPR,sudhakaran2018}. Specifically, when doing cross-validation, we left one person out from the datasets as the test user.

\input{tabActionRecognition.tex}
\input{tabActionRecognitionGTEA71.tex}
\input{tabActionRecognitionGAZEPlus.tex}

\paragraph{Evaluation}
\label{sec:evaluation}

Since our model only uses the temporal structure of the entire video sequence during the training, the evaluation was carried out considering individual video shots and only the first level LSTM. Thus we are using the same given information for testing as previous works on first-person action recognition \cite{simonyan2014two,WangTSN2016,Ma_2016_CVPR,sudhakaran2018}. Additionally, we considered that using the temporal structure information during training to be fair for two reasons. First, our method employs the same information provided in the dataset and even less, considering the GAZE information. Second, it is not stated in the cited literature what information should be used during training. The performance measurements were done using the frame average accuracy score per sequence. In addition, other methods that used the accuracy score at the frame level are indicated on the reported results (Table \ref{tab:actionRecognition}).

\color{Black}

\subsection{Implementation details}
\label{sec:implementation}

\paragraph{Primary regions}
\label{sec:primaryRegions}

Our method estimates the primary region by detecting skin regions and locating wrists points. We trained the skin pixel detector in \cite{Li_2013_ICCV_Workshops} for each video of every dataset using skin masks. For datasets not including the skin mask, a different number of frames was uniformly sampled from each video and the skin contour was annotated using LabelMe \cite{Russell2008}. In the case of the GTEA dataset, we used only the 814 skin masks provided. For the Gaze dataset, an average of 37 frames per video were annotated, giving a total of 625 skin masks. Lastly, 2,230 skin masks were annotated for the GAZE+ dataset, thus having an average of 60 annotated frames per video. The wrist points were detected by using the PAFs network \cite{cao2017realtime}. This network was originally trained using the COCO 2016 keypoints challenge dataset \cite{coco2014}. Specifically, we only used the confidence maps generated for part detection and one scale for the wrist point detection for each dataset.

\paragraph{Secondary regions}
\label{sec:secondaryRegions}
We used the MCG method, pre-trained on the PASCAL 2012 and the BSDS500 datasets, for computing our object proposals. This method generated an average of 1,484 proposals per video frame. 

\paragraph{Data augmentation}
\label{sec:dataAugmentation}


The secondary regions were calculated again for the augmented version of frame $n$ using the MCG algorithm. The effect of the rotation on the MCG was to make new secondary regions that do not overlap with the originals, this augmented the data used by the R*CNN architecture. This process roughly duplicated our training data.

\paragraph{Frame-level modelling}
\label{sec:rStarCNNTraining}

For each dataset, we fine-tuned the fully-connected layers of the R*CNN model on top of a VGG-16 network \cite{Simonyan14c} pretrained on ImageNet. We used stochastic gradient descent (SGD) with a step learning policy and a step decay learning rate of $\gamma=1e^{-1}$ for every 30K iterations. Depending on the dataset, the learning rate $\alpha$ was between $1e^{-4}$ and $2e^{-4}$ with a momentum $\mu=9e^{-1}$ and a batch size of 10. We randomly selected 10 regions from the set of proposal regions, without considering overlapping thresholds with respect to the primary region. The models were trained between 60K and 90K iterations.

\paragraph{Temporal modelling}
\label{sec:rStarCNNplusLSTMTraining}

We trained our hierarchical model by using SGD and the resulting weights of the previous frame-level modeling stage as initialization. The output of the softmax for each frame is provided as input to the hierarchical LSTM model. We first trained the model with $\beta =0$ and then we used the resulting weights as initialization for the training with $\beta >0$. The model with $\beta=0$ was trained by using a step decay learning rate $\gamma=1e^{-1}$ for every 30K iterations. Its learning rate $\alpha$ was between $1e^{-4}$ and $2e^{-4}$, depending on the dataset, a momentum $\mu=9e^{-1}$, and a batch size of 6. The optimization was performed between 30K and 60K iterations. For $\beta >0$, we trained the model between 5 and 8 epochs depending on the dataset by using a step decay learning rate $\gamma\approx9.5e^{-1}$, and starting with a learning rate $\alpha=5e^{-5}$, a momentum $\mu=9e^{-1}$, and a batch size of 6. Since the second LSTM layer captures the sequence of actions across the videos, during test all the video frames are fed sequentially to the network. The prediction at the frame level of the first layer of the LSTM is used to measure the performance. Thus, shot boundary information coming from the second LSTM layer is not needed for testing in our hierarchical model.

\input{tabLevenshteinDistance.tex}

\subsection{Ablation studies}
\label{sec:ablationStudies}

\paragraph{Component Analysis}
\label{sec:componentAnalysis}

On the last rows of Table \ref{tab:actionRecognition} we report performances obtained by using different modules of the proposed architecture. The module that performs spatial reasoning alone determines a consistent improvement with respect to the VGG baseline (13.79 $\%$). The first LSTM layer is responsible for an increase of 8.19 $\%$ of per frame accuracy on average on all datasets. The second LSTM layer is responsible for a further increase of 2.69 $\%$. It is worth to stress that the hierarchical model has a much smaller training set compared to the single-layer LSTM, which explains the relatively small improvement in performance. Moreover, the dissimilarity between the sequences of actions in the datasets explains the difference in performance improvement among them. We measure the similarity as the minimal cost of transforming one sequence into the other using the Levenshtein distance\cite{levenshtein1966binary}, as shown in Table \ref{tab:levenshtein}. The more similar the action sequences are, the higher the accuracy improvement. 

\begin{figure}[!t]
\begin{center}
\includegraphics[width=0.9\columnwidth]{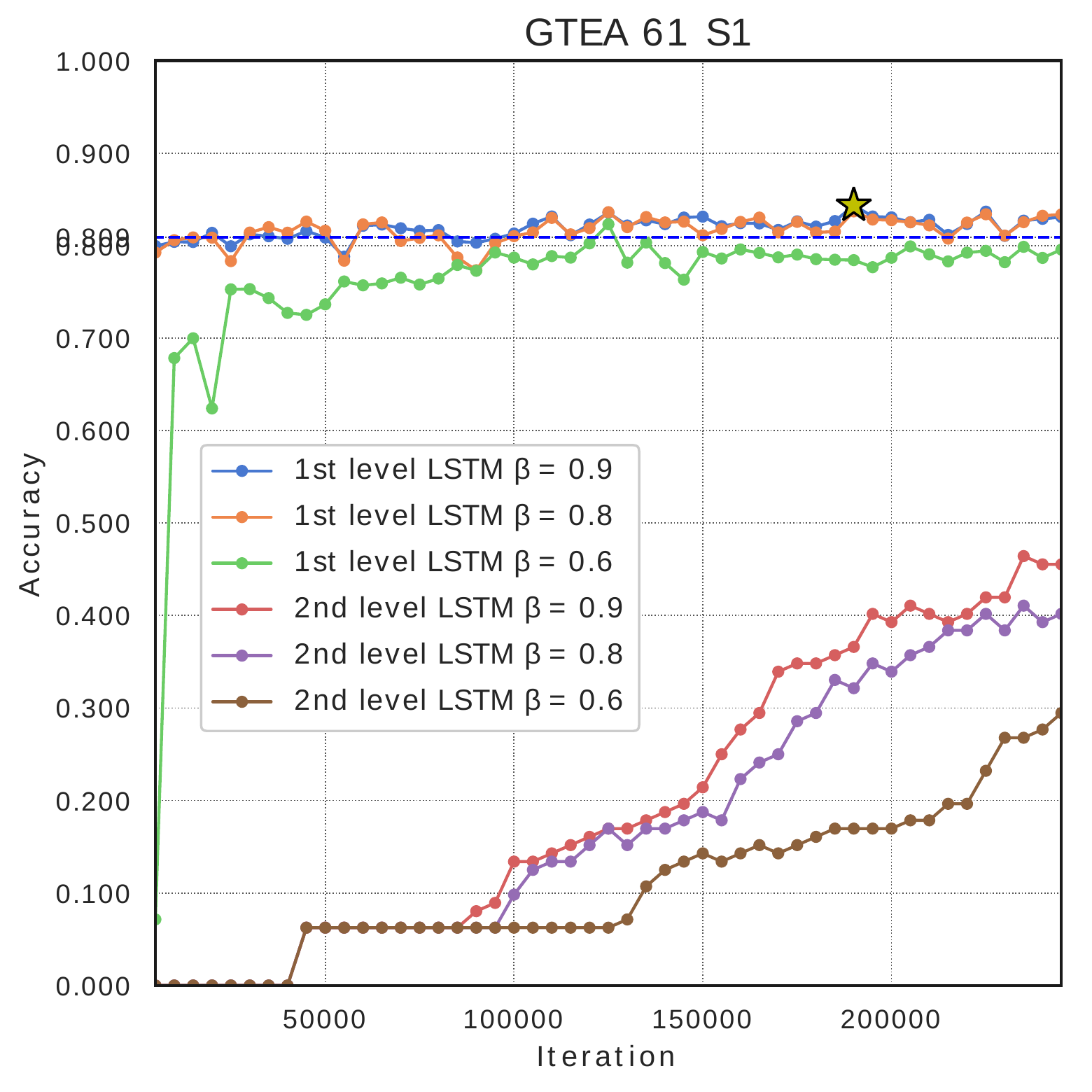}
\caption[]{GTEA 61 test curves for different values of $\beta$. The horizontal dotted blue line indicates the test accuracy obtained for $\beta=0$.}
\label{fig:testCurveGTEA61}
\end{center}
\end{figure}

To better quantify the impact of our frame-level model, we compared its performance to those of the VGG-16 network \cite{Simonyan14c} pretrained on ImageNet, since it serves as the basis for the R*CNN architecture. We observed a considerable improvement of performance of 24.70$\%$ on average. 
The effect of the hierarchical training over the first level LSTM can be seen using different values of $\beta$. First, we trained our model using a $\beta=0$ (i.e. only the first LSTM level) until the optimization process converged. We then performed a grid search varying the learning rate $\alpha$ and the hyperparameter $\beta$ on a single split of each dataset. In particular, we first found a suitable learning rate $\alpha$, and then we looked for the best value of $\beta$ in the set $\{0.5, 0.6, \ldots, 0.9\}$. For example, the effect of different values for $\beta$ can be seen on Fig. \ref{fig:testCurveGTEA61}. Moreover, the plot shows the resulting test curves for the first split of the GTEA 61 dataset. The classification accuracy for the first and second levels are measured per frame and shot, respectively.

\color{Black}

\paragraph{Primary Region Importance}
\label{sec:primaryRegionImportance}

To gain an understanding about the importance of the primary region in our model, we trained the R*CNN architecture with both manually provided regions and with fixed regions covering the low central part of each image. For a single split of the GTEA 71, we achieved $66.11\%$ accuracy when trained with ground truth primary region, $63.80\%$ when trained with our method, $65.05\%$ when trained using the skin as primary region, and $61.93\%$ when trained with a fixed region. Although limited to a single split, these experiments suggest the importance of accurately detecting the primary region. We also report the results of our approach on each evaluated split on Table \ref{tab:splitsGTEA71} and Table \ref{tab:splitsGAZEPLUS}.

	
\paragraph{Visual Data Augmentation}
\label{sec:visualDataAugmentation}

We measured the classification performance of the visual data augmentation by comparing against the original sequence. We considered all the splits from the GAZE+ dataset using the first level LSTM model. The accuracy was obtained using the normal and weighted average prediction for each video shot. The results of this experiment are shown in Table \ref{tab:visualDataAugmentationGAZEPlus}.

\input{tabVisualDataAugmentationGAZEPlus.tex}

\paragraph{Temporal Data Augmentation}
\label{sec:visualDataAugmentation}

In order to understand the effect of temporally augmenting the data sequences, we measure the accuracy performance of our model on the GAZE+ dataset. We trained over all the splits of the dataset using the visually augmented frames. For each split, we trained using as initial weights the best resulting models from the first level LSTM. In addition to calculating the average prediction per shot, we also measured the linearly weighted average prediction.Experimental results are shown in Table \ref{tab:temporalDataAugmentationGAZEPlus}.

\input{tabTemporalDataAugmentationGAZEPlus.tex}

\color{Black}
\subsection{Comparison with the state-of-the-art}
\label{sec:comparisonStateArt}

We tested the proposed approach on five datasets from three well-known benchmarks (GTEA, Gaze, Gaze+) and performed comparisons to five state-of-the-art methods. On average, each competitive method has been tested on 3 datasets. We achieved a gain of $6.11\%$, $5\%$, on the GTEA Gaze, and GTEA Gaze+, respectively. On the GTEA dataset we achieved similar performance to \cite{Ma_2016_CVPR}. On this latter dataset, we observed that often the predictions at frame level within one shot were all incorrect, so that the HLSTM is not able to improve the results for such shots. We stress that with respect to the state-of-the-art, our method does not rely on motion information and shot boundary information is required solely during training. These results demonstrate the effectiveness of reasoning about the spatial arrangement of relevant regions at frame level on the one side, and how important is to exploit temporal structure of videos on the other side 
We observed that, in case of failure, very often either the action verb or the action object are correct. Specifically, a false verb prediction can often be attributed to the lack of motion modeling.

\section{Conclusions}
\label{sec:conclusion}
This paper has introduced a novel approach for egocentric action recognition able to reason
at spatial level about semantically meaningful regions of the image and at temporal level about the sequence of actions being performed. The proposed neural network architecture consists of a CNN module, followed by a HSTM able to capture temporal dependencies within and across shots. A detailed ablation analysis on the various components of our network validate the proposed architecture. The proposed model achieves state-of-the-art results on several standard benchmarks. As future work, we will improve the input to the HLSTM framework, we will exploit motion information and will test our method on the recently introduced EPIC-Kitchens dataset \cite{epicKitchens2018}. 

\section*{Acknowledgements}
This work was partially supported by MINECO/ERDF-EU through the program Ramon y Cajal, projects PID2019-110977GA-I00 and RED2018-102511-T, CONACYT grant 366596. We thank the support of 
NVIDIA Corporation for hardware donation.


\bibliographystyle{IEEEtran}
\bibliography{activity}

\end{document}

%% file: macros.tex
\pdfpageattr{/Group <</S /Transparency /I true /CS /DeviceRGB>>}

\newcolumntype{R}[2]{%
    >{\adjustbox{angle=#1,max width=1.5cm,lap=\width-(#2)}\bgroup}%
    l%
    <{\egroup}%
}

\definecolor{orange}{rgb}{1,0.35,0}
\definecolor{royalblue}{rgb}{0, 0.443137255, 0.737254902}
\definecolor{royalred}{RGB}{155, 28, 49}
\definecolor{trueColor}{RGB}{135, 222, 135}
\definecolor{falseColor}{RGB}{255, 170, 170}



%% file: tabStateArt.tex
\begin{table*}[t]
\caption{State of the art egocentric object-interaction recognition methods and their characteristics}
\resizebox{\textwidth}{!}{%
\centering
\begin{tabular}{r| c c c c c c c  c}
\hline
\multirow{2}{*}{Methods} & \multirow{2}{*}{Hands} & \multirow{2}{*}{Objects} & \multirow{2}{*}{Motion} & \multirow{2}{*}{Gaze} & Spatio-temporal & Spatio-temporal & Spatial reasoning & Temporal reasoning \\
& & & & & feature modelling & reasoning & on hand-objects & on actions  \\
\hline\hline
Fathi et al. (2011) \cite{fathi2011understanding} & \checkmark & \checkmark &  \checkmark &  & \checkmark & & &  \checkmark \\ 
Fathi et al. (2012)\cite{fathi2012learning}  & \checkmark & \checkmark &  \checkmark & \checkmark &  \checkmark & & &  \\ 
Pirsiavash $\&$ Ramanan (2012)\cite{pirsiavash2012detecting}  &    &  \checkmark &  \checkmark & & \checkmark &  & &  \\
McCandless $\&$ Grauman (2013)\cite{mccandless2013object}  &  \checkmark &  \checkmark &  \checkmark & & \checkmark & & & \\
Bambach et al. (2015) \cite{bambach2015lending}  & \checkmark &  & & & & &  \\
Li et al. (2015) \cite{li2015delving}  & \checkmark & \checkmark & \checkmark & \checkmark & & & & \\
Ma et al. \cite{Ma_2016_CVPR} &  \checkmark& \checkmark & \checkmark & & \checkmark & & & \\
Singh et al. \cite{Singh_2016_CVPR} & \checkmark & \checkmark&  \checkmark& & \checkmark &  &  & \\
Sudhakaran $\&$ Lanz (2018)\cite{sudhakaran2018} &  &  \checkmark &  \checkmark& & & \checkmark & &   \\
Baradel et al. (2018)\cite{Baradel_2018_ECCV}  &  & \checkmark & \checkmark  & & &  \checkmark & &  \\
\hline
Ours  &  \checkmark &   \checkmark &  & & & & \checkmark & \checkmark  \\
\hline
\end{tabular}
}
\label{tab:state-of-the-art}
\end{table*}

%% file: tabActionRecognition.tex
\begin{table*}[]
\centering
\begin{tabular}{|r|c|c|c|c|c|c|}
\hline
\textbf{Method} & \textbf{GTEA 61}* & \textbf{GTEA 71} & \textbf{GAZE 25}* & \textbf{GAZE 40}* & \textbf{GAZE+} \\ \hline
Fathi and Rehg $\dagger$~\cite{fathi2013modeling} & 39.7 & - & - & - & -\\  \hline
Li et al. $\dagger$\cite{li2015delving} & 66.8 & 62.1* & 60.9 & 39.60 & 60.50\\ \hline
Two-Streams Network ~\cite{simonyan2014two} & 57.64 & 49.65 & - & - & 58.77 \\ \hline
Temporal Segments Network ~\cite{WangTSN2016} & 67.76 & 67.23 & - & - & 55.25 \\ \hline
Ma et al. \cite{Ma_2016_CVPR} & 75.8 & 73.24 & 62.40 & 43.42 & \textbf{66.40}\\ \hline
Sudhakaran and Lanz \cite{sudhakaran2018} & \textbf{77.59} & \textbf{77} & - & - & 60.13 \\ 
VGG-16 Baseline $\dagger$ & 54.67 & 48.95 & 43.44 & 40.76 & 49.83\\\hline
Ours (frame level) & 68.97 & 64.74 & 56.94 & 47.25 & 52.75\\ \hline
Ours (1 level LSTM) & 69.83 & 71.04 & 63.89 & 49.45 & 58.41\\ \hline
Ours (Hierarchical LSTM) & 70.69 & 72.95 & \textbf{65.28} & \textbf{52.75} & 59.96\\ \hline
\end{tabular}
\caption{Action recognition accuracy on the GTEA, Gaze, and Gaze+ datasets. (Note: * fixed split, $\dagger$ accuracy measured at frame level).}
\label{tab:actionRecognition}
\end{table*}

%% file: tabActionRecognitionGTEA71.tex
\begin{table*}[]
\centering
\begin{tabular}{|r|c|c|c|c|c|}
\hline
Method & Subject 1 & Subject 2 & Subject 3 & Subject 4 & Average\\ \hline
Ours (frame level) & 62.6 & 68.42 & 58.46 & 69.47 & 64.74 \\ \hline 
Ours (1 level LSTM) & 73.28 & 73.68 & 70.0 & 67.18 & 71.04 \\ \hline 
Ours (Hierarchical LSTM) & \textbf{74.05} & \textbf{73.68} & \textbf{73.08} & \textbf{70.99} & \textbf{72.95} \\ \hline
\end{tabular}
\caption{Classification accuracy on each split of the GTEA 71 dataset.}
\label{tab:splitsGTEA71}
\end{table*}

%% file: tabActionRecognitionGAZEPlus.tex
\begin{table*}[]
\centering
\begin{tabular}{|r|c|c|c|c|c|c|c|}
\hline
Method & Subject 1 & Subject 2 & Subject 3 & Subject 4 & Subject 5 & Subject 6 & Average\\ \hline
Ours (frame level) & 49.8 & 58.05 & 46.11 & 60.66 & 51.02 & 50.83 & 52.75 \\ \hline 
Ours (1 level LSTM) & 56.55 & \textbf{65.77} & 47.58 & 63.16 & 63.27 & 54.17 & 58.41 \\ \hline 
Ours (Hierarchical LSTM) & \textbf{58.53} & 64.43 & \textbf{51.37} & \textbf{64.54} & \textbf{64.63} & \textbf{56.25} & \textbf{59.96} \\ \hline 
\end{tabular}
\caption{Classification accuracy on each split of the GAZE+ dataset.}
\label{tab:splitsGAZEPLUS}
\end{table*}

%% file: tabLevenshteinDistance.tex
\begin{table}[]
\centering
\resizebox{\columnwidth}{!}{%
\begin{tabular}{|c|c|c|c|c|c|c|}
\hline
& GTEA 61 & GTEA 71 & GAZE 25 & GAZE 40 & GAZE+ \\ \hline
Avg. Levenshtein & \multirow{2}{*}{\large2.36} & \multirow{2}{*}{\large2.57} & \multirow{2}{*}{\large15.12} & \multirow{2}{*}{\large19.40} & \multirow{2}{*}{\large50.40} \\
distance & & & & & \\ \hline
\end{tabular}
}
\caption{Similarity between sequences of actions in videos for each dataset. The closer the value to zero the more similar.}
\label{tab:levenshtein}
\end{table}

%% file: tabVisualDataAugmentationGAZEPlus.tex
\begin{table*}[]
\centering
\begin{tabular}{|c|c|c|c|c|c|c|c|c|}
\hline
\multirow{2}{*}{Shot Evaluation} & \multirow{2}{*}{Sequence} & \multicolumn{6}{c|}{Splits} & \multirow{2}{*}{Average}\\ \cline{3-8}
 & & Subject 1 & Subject 2 & Subject 3 & Subject 4 & Subject 5 & Subject 6 & \\ \hline \hline
\multirow{2}{*}{Avg. Prediction} & Original &  55.56 & 64.09 & 45.89 & 62.88 & 61.9 & \textbf{56.67} & 57.83 \\ \cline{2-9}
 & Augmented & \textbf{56.55} & \textbf{65.77} & \textbf{47.58} & \textbf{63.16} & \textbf{63.27} & 54.17 & \textbf{58.41} \\ \hline \hline
Weighted        &  Original &          54.56 & \textbf{63.76} &         47.79  &         62.88  & \textbf{62.59} & \textbf{57.08} & \textbf{58.11} \\ \cline{2-9}
Avg. Prediction & Augmented & \textbf{56.94} & \textbf{63.76} & \textbf{49.47} & \textbf{63.16} & 60.54 & 54.58 & 58.08 \\ \hline 
\end{tabular}
\caption{Classification performance comparison using visual data augmentation on the GAZE+ dataset for 44 categories.}
\label{tab:visualDataAugmentationGAZEPlus}
\end{table*}

%% file: tabTemporalDataAugmentationGAZEPlus.tex
\begin{table*}[]
\centering
\begin{tabular}{|c|c|c|c|c|c|c|c|c|}
\hline
\multirow{2}{*}{Shot Evaluation} & \multirow{2}{*}{Sequence} & \multicolumn{6}{c|}{Splits} & \multirow{2}{*}{Average}\\ \cline{3-8}
 & & Subject 1 & Subject 2 & Subject 3 & Subject 4 & Subject 5 & Subject 6 & \\ \hline \hline
\multirow{2}{*}{Avg. Prediction} & Original & %
\textbf{58.53} & 64.43          & \textbf{51.37} & \textbf{64.54} & \textbf{64.63} & \textbf{56.25} & \textbf{59.96} \\ \cline{2-9}
 & Augmented & %
58.33          & \textbf{65.44} & \textbf{51.37} & 63.99          & \textbf{64.63} & 55             & 59.79 \\ \hline \hline
Weighted &  Original & %
\textbf{58.13} & 63.42          & \textbf{52.21} & 63.44          & 62.59          & \textbf{57.08} & 59.48 \\ \cline{2-9}
Avg. Prediction & Augmented & %
        57.74  & \textbf{64.09} & 52     & \textbf{65.10} & \textbf{63.26} & 56.67          & \textbf{59.81} \\ \hline
\end{tabular}
\caption{Classification performance comparison using temporal data augmentation on the GAZE+ dataset for 44 categories.}
\label{tab:temporalDataAugmentationGAZEPlus}
\end{table*}